\documentclass[a4paper,twoside]{article}
\usepackage{epsfig}
\usepackage{subcaption}
\usepackage{calc}
\usepackage{amssymb}\usepackage{amstext}\usepackage{amsmath}\usepackage{amsthm}
\usepackage{multicol}
\usepackage{pslatex}
\usepackage{apalike}
\usepackage{algorithm2e}
\usepackage[bottom]{footmisc}
\usepackage{SCITEPRESS}
\usepackage[hidelinks]{hyperref}

\newif\ifblind \blindfalse

\title{Continuous-Time Set-Transformers for Asynchronous WiFi-IMU Indoor Localization}

\ifblind

\else
  \author{\authorname{Mohamed Bachar\orcidAuthor{0009-0001-6222-3096}, Ilyass Abouelaziz\orcidAuthor{0000-0001-7208-2509} and Yuehua Ding\orcidAuthor{0000-0003-3100-2204}}
    \affiliation{CESI, CESI LINEACT, UR 7527, France}
    \email{\{mbachar, iabouelaziz, yding\}@cesi.fr}}
\fi

\abstract{
Indoor localization commonly combines WiFi received signal strength indicator (RSSI) fingerprints with inertial measurements. However, these sensor streams are sampled asynchronously at different rates; therefore, observations rarely align in time and may be missing or stale. Most existing methods resample the data onto a shared clock and process each modality using a separate branch. We instead present a continuous-time transformer that represents observations as an unordered set of timestamped tokens. Each token encodes its measurement, modality, and elapsed time since acquisition. A single attention block jointly fuses information across modalities and time, avoiding both resampling and modality-specific branches while remaining invariant to token order. During training, we randomly mask entire modalities and individual observations, enabling the model to operate under missing or outdated inputs. We validate the modality-specific encoders on their respective benchmarks and evaluate the fused model on both simulated and real datasets, including a cross-session dataset collected on unseen days. The proposed approach improves upon an open WiFi fingerprinting baseline under cross-session drift, remains competitive with learned fusion and dead-reckoning baselines, and degrades gracefully as observations become missing or stale. These results suggest that a single continuous-time transformer provides a practical framework for indoor localization from asynchronous sensors.
}

\begin{document}
\onecolumn \maketitle \normalsize \setcounter{footnote}{0} \vfill

\section{\uppercase{Introduction}}

\noindent Many agents that move through buildings must know where they are. A pedestrian
following turn-by-turn directions on a smartphone, a mobile robot running
warehouse logistics or facility inspection, and a tracked asset reporting its
location all depend on a position estimate. Outdoors that estimate comes from
GPS, but satellite signals scarcely penetrate walls. Consequently, an indoor device must
infer its position from the sensors it already carries. The inference is
error prone because no effective global reference is available indoors: the
measurements are noisy, and the environment shifts as people and objects move.
Accuracy is bounded further by the sensing hardware and by the site survey that
calibrates it. An estimate that holds under one set of conditions can degrade
sharply once any of them changes, so reliable indoor self-localization remains an
open problem.

Two sensors are carried by nearly all of these devices: a WiFi radio and an
inertial measurement unit (IMU). A WiFi scan gives a sparse absolute reference to
known access points. The IMU reports dense relative motion at a significantly higher rate.
The two streams are complementary, yet heterogeneous and sampled asynchronously.
No common clock aligns their arrivals, and either stream may become unavailable or arrive
stale during operation. Fusing these imperfect, asynchronous signals into one
position estimate is the core problem this work targets.

In this paper, we propose a continuous-time transformer (Figure~\ref{fig:pipeline})
that fuses asynchronous WiFi and IMU observations into a single $(x, y)$ estimate
within one model that remains accurate when a sensor is missing or stale. The
approach is not tied to a particular platform. It needs only WiFi scans and
inertial windows, the same inputs a phone, a robot, or a wearable already produces,
so it applies to any device that carries these two sensors. We validate it on real
cross-session data.

The remainder of this paper is organized as follows: Section~\ref{sec:related}
reviews related work, Section~\ref{sec:problem} states the problem,
Section~\ref{sec:method} develops the method, Section~\ref{sec:experiments}
describes the experimental setup, Section~\ref{sec:results} reports the results,
and Section~\ref{sec:conclusion} concludes.

\section{\uppercase{Related Work and Motivation}}\label{sec:related}

\noindent WiFi received signal strength indicator (RSSI) fingerprints recover absolute
indoor position by matching a live scan to a labeled survey~\cite{bahl2000radar}.
On UJIIndoorLoc~\cite{torressospedra2014ujiindoorloc}, a 1-NN reference method
reports a 7.9~m error. Such fingerprinting stays single-modality, encodes neither
elapsed time nor inertial motion, and drifts as environment and hardware change,
which motivates pairing WiFi with a dense motion modality.

Learned inertial navigation provides dense relative estimates.
RoNIN~\cite{yan2019ronin} is the standard benchmark, and attention entered the
field through iMoT~\cite{nguyen2025imot}, which uses accelerometer and gyroscope
channels. Yet the sampling rate is fixed by the window length or token dimension,
and inertial tracking drifts due to the absence of absolute reference; combining WiFi's
absolute reference with inertial motion is the remedy.

The fusion of these two heterogeneous streams is well studied, but mostly through
a separate branch per modality, merged by a classical filter or late
concatenation. WIO-EKF, evaluated across a ten-day session
gap~\cite{zhou2024wioekf}, keeps each sensor in its own EKF branch. Deep variants
concatenate recurrent streams or substitute a null vector when WiFi is
absent~\cite{wei2021sensorfusion}, and where attention appears it sits inside a
branch rather than as the fusion itself. Two limits recur: streams are resampled
onto a common grid before fusion, and robustness to a missing sensor is hand-set
or filter-based rather than learned. Evaluation also tends to use random or
cross-user splits, leaving generalization across sessions lightly tested.

The ingredients for a different representation already exist, scattered across
domains. The Transformer's positional encoding assumes equidistant
positions~\cite{vaswani2017transformer}, which breaks for multi-rate sensors,
whereas set models are permutation-invariant over unordered
inputs~\cite{lee2019settransformer}. Dropping an entire modality at training time,
the precedent for robustness to an absent channel, originates with
ModDrop~\cite{neverova2014moddrop}. The learned embedding of real-valued time
builds on Time2Vec and mTAN~\cite{kazemi2019time2vec,shukla2021mtan}. SeFT and
STraTS realize a single continuous-time set transformer without an ODE solver, but
for clinical data~\cite{horn2020seft,tipirneni2022strats}. Two differences set the
proposed method apart from STraTS, beyond the domain shift. It adds an encoder per
modality that maps heterogeneous WiFi RSSI scans and inertial windows into the
shared token, where STraTS assumes scalar clinical observations. It also trains
with modality and instant dropout for robustness across sessions, which STraTS
does not address. The closest localisation
assemblies remain partial: AFT-VO fuses asynchronous sources in one transformer
but is camera-only and bins time~\cite{kaygusuz2022aftvo}, while ContiFormer puts
an ODE solver inside attention at high cost~\cite{chen2023contiformer}.

Three properties separate these assemblies, and each holds only a subset.
P1 is continuous-time tokens that avoid resampling and an ODE solver; P2 is one
permutation-invariant set transformer doing cross-modal and cross-time fusion in
a single self-attention block; P3 is learned robustness to missing or stale
sensors, validated across sessions. Table~\ref{tab:properties} scores the methods
discussed above against the three.

\begin{table}[tb]
\caption{Methods scored against the three properties. P1: continuous-time tokens
  without resampling or an ODE solver. P2: one permutation-invariant set
  transformer for cross-modal and cross-time fusion in a single self-attention
  block. P3: learned robustness to missing or stale sensors, validated across
  sessions. \checkmark\ holds, $\sim$ partial, -- absent or not applicable.}
\label{tab:properties}
\centering
\begin{tabular}{|l|l|c|c|c|}\hline
Method & Domain & P1 & P2 & P3 \\ \hline
iMoT              & IMU       & --        & --        & --        \\ \hline 
AFT-VO         & camera    & --        & $\sim$    & --        \\ \hline 
ContiFormer & general   & $\sim$    & --        & --        \\ \hline 
SeFT / STraTS & clinical & \checkmark & \checkmark & -- \\ \hline 
WIO-EKF          & WiFi+IMU  & --        & --        & $\sim$    \\ \hline 
Deep concat fusion & WiFi+IMU & --     & --        & $\sim$    \\ \hline 
Ours                                   & WiFi+IMU  & \checkmark & \checkmark & \checkmark \\ \hline 
\end{tabular}
\end{table}

The proposed method completes the combination for asynchronous WiFi--IMU
localization. Each observation carries a learned sinusoidal encoding of its
real-valued elapsed $\Delta t$, a lightweight alternative to resampling and
ODE-in-attention. One set transformer performs cross-modal and cross-time fusion
in a single self-attention block. Modality and instant dropout extend the earlier
whole-modality scheme to the (modality, time) token set, giving graceful
degradation under missing and stale sensors that evaluation across sessions
confirms.

\section{\uppercase{Problem Statement}}\label{sec:problem}

\noindent
The objective of self-localization is for an agent to estimate its own position from heterogeneous data, asynchronously sampled by different sensors. For heterogeneous data fusion, every sensor reading is represented in a homogeneous form, as follows:
\[(s_i,\, m_i,\, t_i,\, a_i)\]
where $s_i$ is the raw sample, with samples
drawn from different spaces,
\begin{equation}\label{eq:spaces}
  s_i \in
  \begin{cases}
    \mathbb{R}^{n_{\mathrm{ap}}}, & \text{if~} m_i=\text{wifi},\\[2pt]
    \mathbb{R}^{T\times F},      & \text{if~} m_i=\text{imu},
  \end{cases}
\end{equation}
an RSSI vector over $n_{\mathrm{ap}}$ access points for WiFi, and a $T$-step
window of $F$ inertial channels for the IMU. $m_i \in \mathcal{M}$ with $\mathcal{M}=\{\text{wifi},\text{imu}\}$ represents its modality
(sensor type), $t_i$ is the acquisition time, and $a_i \in \{0,1\}$ is an availability flag ('1' stands for 'available'). 

A WiFi scan gives an absolute place reference but arrives sparsely, whereas
an IMU reports dense relative motion at a significantly higher rate. As the IMU reporting rate may largely exceed the WiFi sampling rate, one can hardly have samples of both modalities at a given instant. A common time grid would discard the fast stream or pad the slow
one. The input is therefore an unordered set of observations, not a
synchronized sequence.

Upon a localization query at time instant $t_q$, the model groups its $N$ recent observations in a set $\mathcal{O}'$, as follows:
\begin{equation}\label{eq:obsset-gen}
  \mathcal{O}' = \big\{\, (s_i,\, m_i,\, t_i,\, a_i) \,\big\}_{i=1}^{N}.
\end{equation}
The general problem is finding a method $f$ capable of mapping the observations $\mathcal{O}'$ to a planar position $(\hat{x},\hat{y})$, which is mathematically formulated as follows:
\begin{equation}\label{eq:objective-gen}
  \begin{aligned}
    \min_{f}\ \ & \big\lVert (\hat{x},\hat{y}) - (x,y) \big\rVert^2 \\
    \text{s.t.}\ \ & (\hat{x},\hat{y}) = f\big(\mathcal{O}',\, t_q\big),
  \end{aligned}
\end{equation}
where $(\hat{x},\hat{y})$ is the prediction and $(x,y)$ the ground truth.

Two conditions make the problem hard. First, deployment is across sessions:
$f$ is fit on one set of recording sessions and applied to others, the
setting that carries our main claim. Second, the input is incomplete, since
at a query a modality may be missing, with no available sample, or stale,
with only an old one. The method of Section~\ref{sec:method} must absorb
both of these.

\section{\uppercase{Proposed Method}}\label{sec:method}

\begin{figure*}[!t]\centering
  \includegraphics[width=0.9\textwidth]{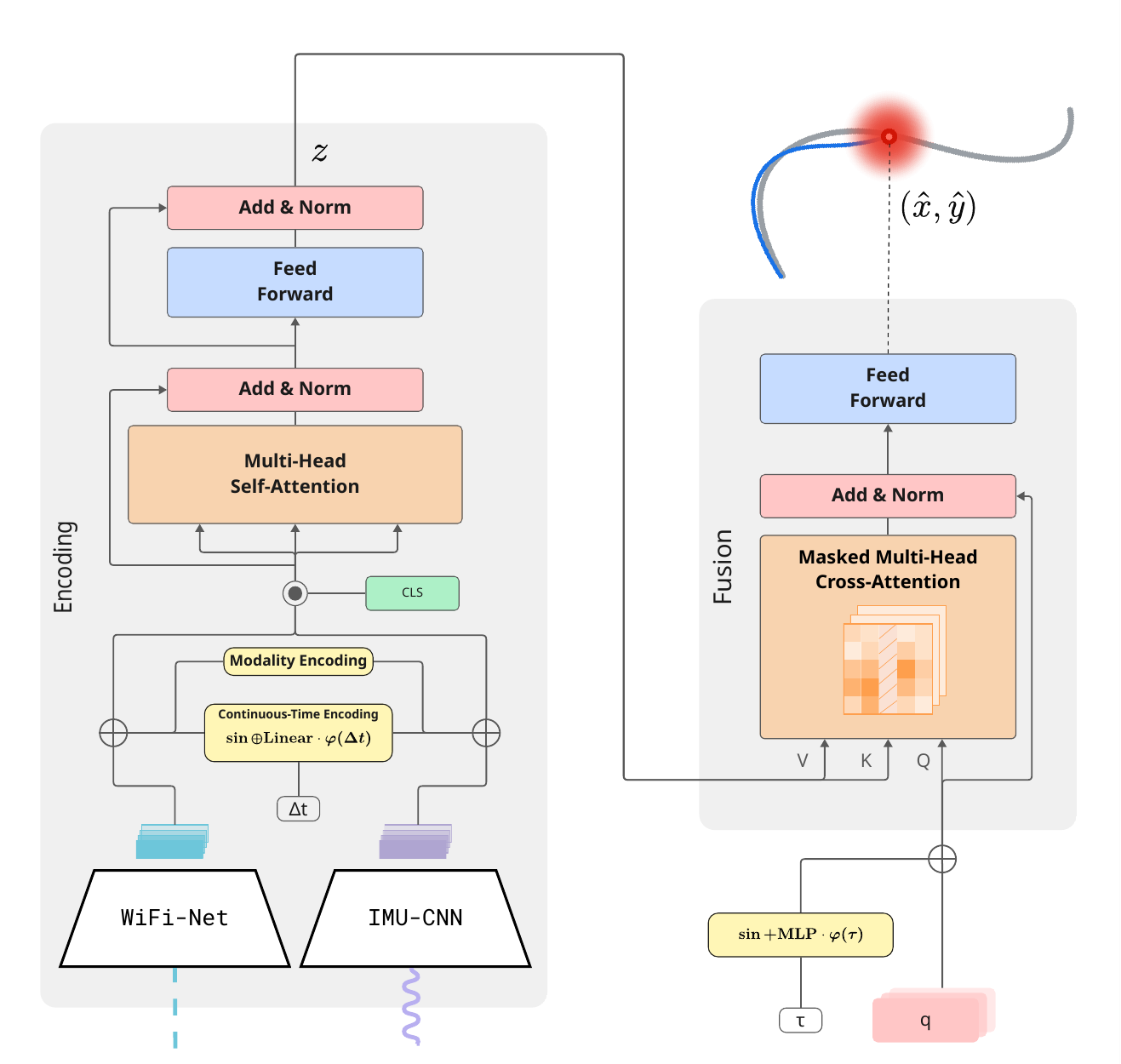}
  \caption{Overview of the proposed continuous-time transformer: each
    observation becomes one token, a single self-attention block fuses the set,
    and a learnable position query reads out $(\hat{x},\hat{y})$.}
  \label{fig:pipeline}
\end{figure*}

\noindent In this section, a single continuous-time set transformer is proposed to regress from the asynchronous WiFi and IMU streams to a planar position, with neither branch per modality nor resampling onto a common clock.
To this end, the problem is reformulated and analyzed in the upcoming subsections.
\subsection{Problem Reformulation and Functional Decomposition}
\noindent
In general, the absolute time instant does not influence the localization results in a relatively stable environment, but the timeliness of the data does. For localization query at instant $t_q$, the timeliness of observations $\mathcal{O}'$ is measured by the signed elapsed time
$\Delta t_i = t_i - t_q$, non-positive for a past observation. This removes $t_q$
from the inputs and casts the observations as an unordered set
\begin{equation}\label{eq:obsset}
  \mathcal{O} = \big\{\, (s_i,\, m_i,\, \Delta t_i,\, a_i) \,\big\}_{i=1}^{N},
\end{equation}
holding $K$ recent instants per modality, so $N=KM$ slots are addressable and a
slot with $a_i=0$ contributes no token. One should note that the estimate of the position depends on the sample values in set $ \mathcal{O} $ rather than the order of the samples, so the map is invariant to any
permutation $\pi$ of the set,
\begin{equation}\label{eq:perm}
  f(\pi\!\cdot\!\mathcal{O}) = f(\mathcal{O}).
\end{equation}
Because the elements of $\mathcal{O}$ differ in type, dimension and timing, $f$ is decomposed into an \emph{encoding} unit $g$ and a set-level
\emph{fusion} unit $h$, which is described as:
\begin{equation}\label{eq:decomp}
  \begin{gathered}
    f = h \circ g,\\
    g:(s_i,m_i,\Delta t_i)\mapsto \mathbf{z}_i\in\mathbb{R}^{D},\\
    h:\{\mathbf{z}_i\}\mapsto(\hat x,\hat y).
  \end{gathered}
\end{equation}

The encoding and fusion process in (\ref{eq:decomp}) can be illustrated by Figure~\ref{fig:pipeline}. Each sensor sample, whatever its rate, becomes one token of a shared latent
space, and a single self-attention block performs the entire fusion.
Figure~\ref{fig:pipeline} traces this end to end: an encoder for each modality
realizes $g$, mapping each observation to one universal token; the attention
block and a learnable position query realize $h$, which contextualizes the set
and reads out $(\hat{x},\hat{y})$.

Modality identity is carried by an additive embedding, rather than by token order; therefore, a
modality added or removed by inserting or deleting tokens brings no change to the
trunk. Tolerance to missing and stale sensors is then a property of training, without changing the architecture.

\subsection{Universal Tokenization and Encoders for Each Modality}
\label{subsec:tok}

\noindent
This section realizes the encoding $g$ of Equation~\eqref{eq:decomp}, mapping each
element of the observation set $\mathcal{O}$ to a common latent space; the fusion
$h$ that delivers the permutation invariance of Equation~\eqref{eq:perm} follows in
Section~\ref{subsec:fusion}.

The encoding $g$ sends every observation to one $D$-dimensional token holding the
measurement, its modality, and its elapsed time relative to the query,
\begin{equation}\label{eq:token}
  \mathbf{z}_i = \mathrm{enc}_{m_i}(s_i)
               + \mathbf{e}_{m_i}
               + \phi(\Delta t_i).
\end{equation}
The first term is the output of the encoder for that modality; the second a
learnable modality embedding $\mathbf{e}_m \in \mathbb{R}^{D}$, carrying sensor identity additively so it survives
permutation; the third the continuous-time encoding of $\Delta t_i$.

\paragraph{Continuous-time encoding.}
The encoding $\phi$ admits sensors of widely different rates into one model
without resampling. Unlike the discrete positional encoding of sequence
transformers, $\phi$ is a function of the real-valued $\Delta t$ in seconds. It
evaluates $n$ sinusoids whose angular frequencies $\boldsymbol{\omega} =
2\pi/\boldsymbol{\tau}$ derive from periods $\boldsymbol{\tau}$ spaced
geometrically over $[\tau_{\min},\tau_{\max}]$. Taking the sine and cosine of
each gives $2n$ features, which a learned matrix
$\mathbf{W}_\phi \in \mathbb{R}^{D \times 2n}$ projects to $D$ dimensions,
\begin{equation}\label{eq:time}
  \phi(\Delta t) = \mathbf{W}_\phi
    \big[\, \sin(\boldsymbol{\omega}\,\Delta t) \,\|\,
            \cos(\boldsymbol{\omega}\,\Delta t) \,\big].
\end{equation}
Under the geometric spacing, observation times close together receive similar
codes while distant times stay separable, across periods from $0.05$~s to
$120$~s using $n=32$ sinusoids. Reading the continuous $\Delta t$ rather than a
discrete index places every token on one shared time axis regardless of when it
arrived.

\begin{figure}[!hb]\centering
  \includegraphics[width=1\columnwidth]{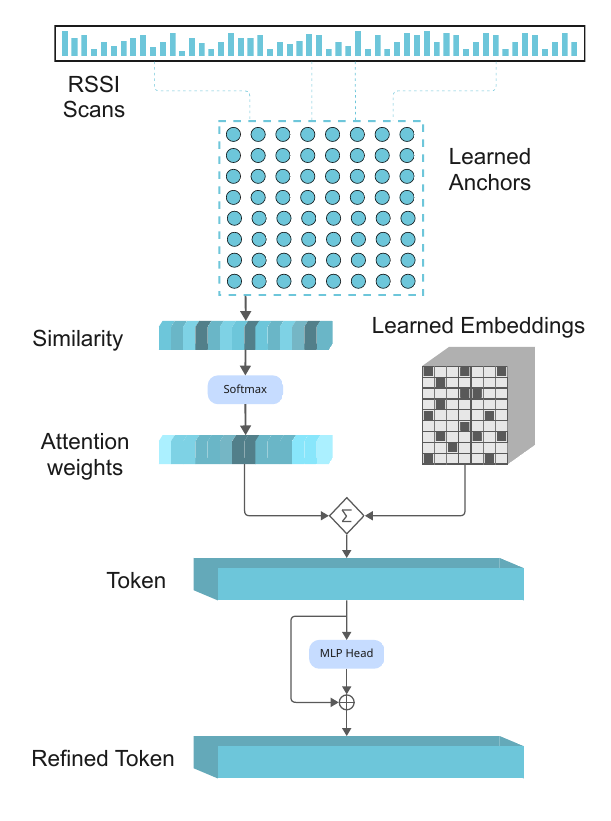}
  \caption{WiFi-Net encoder. An RSSI scan is scored against $N_a$ learned
    reference points; a softmax over the similarities gives attention weights that
    aggregate the learned embeddings of those reference points into a token,
    refined by a residual MLP head.}
  \label{fig:wifinet}
\end{figure}

\paragraph{WiFi-Net.}
WiFi-Net encodes the modality that fixes absolute place. It treats a single RSSI
scan as a query against a learned set of $N_a$ reference points, producing a
distribution that discriminates place rather than a raw radio map. The scan
$s \in \mathbb{R}^{n_{\mathrm{ap}}}$ (Equation~\eqref{eq:spaces}) is scored against
the matrix of reference points $\mathbf{A} \in \mathbb{R}^{N_a \times n_{\mathrm{ap}}}$,
and a softmax with learned temperature $\gamma$ yields a soft assignment
$\mathbf{w}$ over the reference points,
\begin{equation}\label{eq:wifi_assign}
  \mathbf{w} = \mathrm{softmax}\!\left( \frac{\mathbf{A}\, s}{\gamma} \right)
            \in \Delta^{N_a-1},
\end{equation}
a point on the $(N_a{-}1)$-simplex. The weights $\mathbf{w}$ aggregate the
embeddings of the reference points $\mathbf{E} \in \mathbb{R}^{N_a \times D}$ into
a pooled vector, which a residual MLP with layer normalization then refines into
the token,
\begin{equation}\label{eq:wifi_token}
  \mathrm{enc}_{\mathrm{wifi}}(s) = \mathbf{t} + \mathrm{MLP}(\mathbf{t}),
  \qquad
  \mathbf{t} = \sum_{j=1}^{N_a} w_j\, \mathbf{E}_j = \mathbf{w}^{\!\top}\mathbf{E}.
\end{equation}
The reference points are learned, not tied to physical access points, so the
encoder positions them where they best discriminate place. We use $N_a=64$
reference points (Figure~\ref{fig:wifinet}).

\paragraph{IMU-CNN.}
IMU-CNN summarizes dense motion over a short temporal window, supplying the
relative motion that complements WiFi's absolute reference. The useful inertial
signature is a \emph{local} pattern in time, a brief acceleration or turn whose
contribution to position should not depend on where in the window it falls. The
encoder reads a fixed window $\mathbf{X} \in \mathbb{R}^{T \times F}$, the IMU
sample of Equation~\eqref{eq:spaces}, with $\mathbf{X}_{t,f}$ the $f$-th inertial
channel at step $t$.

The encoder is a stack of three 1D-convolutional blocks. Each block convolves a
learned filter bank along the time axis, then applies batch normalization, a
GELU nonlinearity $\sigma$ and dropout,
\begin{equation}\label{eq:imu_conv}
  \mathbf{U}^{(\ell)} = \sigma\!\big(
    \mathrm{BN}( \mathbf{W}^{(\ell)} * \mathbf{U}^{(\ell-1)} ) \big),
  \qquad \ell = 1, 2, 3,
\end{equation}
where $*$ denotes 1D convolution along time, $\mathbf{W}^{(\ell)}$ is the learned
filter bank of block $\ell$, and $\mathbf{U}^{(0)} = \mathbf{X}^{\!\top}$. Sharing
filter weights across time makes each block shift-equivariant, so a motion pattern
produces the same response wherever it falls. The three blocks raise the channel
count $F \!\to\! C_1 \!\to\! C_2 \!\to\! C_3$ and compose local responses into
deeper motion features $\mathbf{U} \equiv \mathbf{U}^{(3)} \in \mathbb{R}^{C_3
\times T'}$.

\begin{figure}[!hb]\centering
  \includegraphics[width=1\columnwidth]{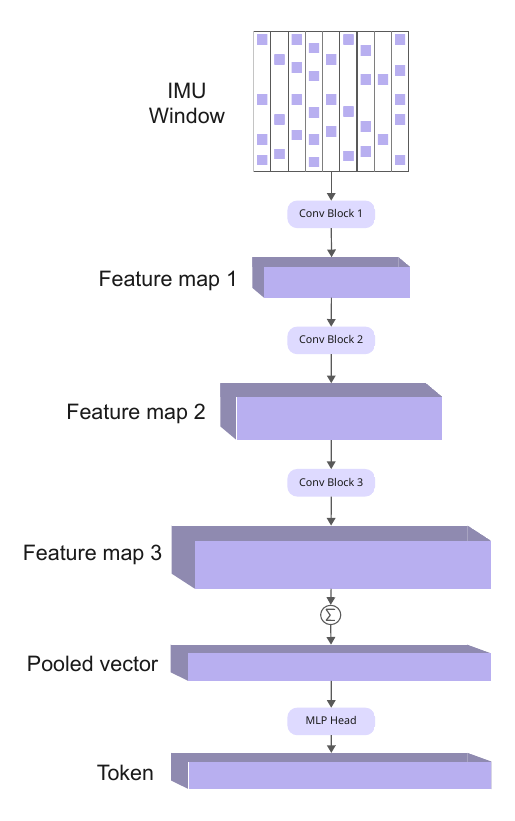}
  \caption{IMU-CNN encoder. A $T\times F$ inertial window passes through three
    convolutional blocks; global average pooling collapses the final feature
    map over time into a pooled vector that an MLP head maps to the token.}
  \label{fig:imucnn}
\end{figure}

Global average pooling over time then collapses the window, averaging each
channel's response across all positions,
\begin{equation}\label{eq:imu_gap}
  \bar{\mathbf{u}} = \frac{1}{T'} \sum_{t=1}^{T'} \mathbf{U}_{:,t}
    \;\in\; \mathbb{R}^{C_3}.
\end{equation}
Pooling turns the blocks' shift-equivariance into shift-invariance: any temporal
shift of a feature leaves $\bar{\mathbf{u}}$ unchanged. A head of a linear layer
$\mathbf{W}_o$ with layer normalization sends this summary to the $D$-dimensional
token,
\begin{equation}\label{eq:imu}
  \mathrm{enc}_{\mathrm{imu}}(\mathbf{X})
    = \mathrm{LN}\!\big( \mathbf{W}_o\, \bar{\mathbf{u}} \big).
\end{equation}
The window holds $T=32$ steps of $F=9$ inertial features (triaxial acceleration,
triaxial angular rate, and roll, pitch, yaw), with channels
$(C_1,C_2,C_3)=(32,64,128)$; the learned filters capture dead reckoning over short
intervals (Figure~\ref{fig:imucnn}). Both encoders emit the same $D=128$ token
width, so each enters the fusion block without wiring specific to its modality.
IMU-CNN holds about $0.05$~M parameters.

\subsection{Set-Transformer Fusion and Position Readout}
\label{subsec:fusion}

\noindent
This section realizes the fusion $h$ of Equation~\eqref{eq:decomp}. The tokens of
Equation~\eqref{eq:token}, one per available observation, form the set the fusion
block aggregates. Self-attention is permutation equivariant by construction. This
is exactly the aggregation that the token encoding of Section~\ref{subsec:tok} is
built for. Once the set spans both modalities and $K$ instants, the one operation
performs cross-modal and cross-time fusion at once: a WiFi token attends to an IMU
token from a different instant exactly as to one from its own. The block stacks
$L$ pre-norm transformer encoder layers with $H$ attention heads, a feed-forward
width of $4D$ and a GELU nonlinearity. Each layer applies multi-head self-attention
(MHSA) and a position-wise feed-forward network (FFN) with residual connections,
\begin{equation}\label{eq:layer}
  \begin{aligned}
    \hat{\mathbf{Z}} &= \mathbf{Z}
      + \mathrm{MHSA}\big(\mathrm{LN}(\mathbf{Z}),\, \mathbf{B}\big), \\
    \mathbf{Z}' &= \hat{\mathbf{Z}}
      + \mathrm{FFN}\big(\mathrm{LN}(\hat{\mathbf{Z}})\big),
  \end{aligned}
\end{equation}
for a token stack $\mathbf{Z}$ and an additive padding bias $\mathbf{B}$.
Multi-head self-attention runs $H$ scaled-dot-product heads in parallel,
\begin{equation}\label{eq:mhsa}
  \mathrm{head}_h = \mathrm{softmax}\!\left(
    \frac{\mathbf{Q}_h \mathbf{K}_h^{\!\top}}{\sqrt{d_h}} + \mathbf{B}
  \right)\mathbf{V}_h,
  \quad d_h = \tfrac{D}{H},
\end{equation}
and concatenates them through an output projection. The padding bias $\mathbf{B}$
assigns $-\infty$ to every token that is absent ($a_i=0$) or dropped, so it
receives no attention mass. The set size $N$ then enters only through which tokens
are live, which is how the missing and stale conditions of
Section~\ref{sec:problem} act on the model. A learnable classification (CLS) token
is prepended and never masked, so every attention row keeps at least one live key.

The readout maps the contextualized set to a position through a single learnable
position query $\mathbf{q}$ (Figure~\ref{fig:pipeline}). The query cross-attends
to the contextualized set, with the CLS token retained in its key set, and a
two-layer MLP maps the attended vector to the predicted position
$\hat{\mathbf{p}} = (\hat{x},\hat{y})$,
\begin{equation}\label{eq:readout}
  \hat{\mathbf{p}} = \mathrm{MLP}\!\Big(
    \mathrm{CrossAttn}\big(\mathbf{q} + \phi(\tau),\; [\mathbf{c}\,\|\,\mathbf{Z}'],\; \mathbf{B}\big)
  \Big),
\end{equation}
where $\mathbf{Z}'$ is the contextualized token set, $\mathbf{c}$ the
contextualized CLS token, $\mathbf{q}$ the learnable query, and $\phi(\tau)$ the
continuous-time encoding of Equation~\eqref{eq:time} evaluated at the query offset
$\tau$, so position can be read at a chosen time. Collapsing the
permutation-equivariant set to a single output makes the full map $f$
permutation-invariant, as Equation~\eqref{eq:perm} requires.

Equation~\eqref{eq:objective-gen} poses the objective as squared error. Training
minimizes a Huber surrogate $\mathcal{H}_\delta$, quadratic for small residuals and
linear beyond a threshold $\delta$, which limits the pull of outlier positions,
\begin{equation}\label{eq:loss}
  \begin{gathered}
    \mathcal{L} = \frac{1}{N_s}\sum_{i=1}^{N_s}
      \mathcal{H}_\delta\big(\hat{\mathbf{p}}_i - \mathbf{p}_i\big),\\
    \mathcal{H}_\delta(r) =
    \begin{cases}
      \tfrac12 \lVert r\rVert^2, & \lVert r\rVert \le \delta,\\[2pt]
      \delta\big(\lVert r\rVert - \tfrac12\delta\big), & \text{otherwise,}
    \end{cases}
  \end{gathered}
\end{equation}
with $\mathbf{p}_i$ the ground truth position. The optimizer and the threshold
$\delta$ are given in Section~\ref{sec:experiments}, along with the modality and
instant dropout that confer robustness to missing and stale sensors.
Algorithm~\ref{alg:forward} assembles the encoding $g$, the fusion $h$ and the
loss into one pass.

\begin{algorithm}[!h]
\caption{Forward pass and training loss. The encoding $g$ builds the token set;
  the fusion $h$ contextualizes it and reads out the position, so
  $\hat{\mathbf{p}} = h(g(\mathcal{O})) = f(\mathcal{O})$.}\label{alg:forward}
\SetAlgoLined
\DontPrintSemicolon
\KwIn{observation set $\mathcal{O} = \{(s_i, m_i, \Delta t_i, a_i)\}$; query
  offset $\tau$; ground truth $\mathbf{p}$ (training only)}
\KwOut{prediction $\hat{\mathbf{p}} = (\hat{x},\hat{y})$; training loss $\mathcal{L}$}
\For{$i : a_i = 1$}{
  $\mathbf{z}_i \leftarrow \mathrm{enc}_{m_i}(s_i) + \mathbf{e}_{m_i}
    + \phi(\Delta t_i)$ \tcp*{encoding $g$, Eq.~\eqref{eq:token}}
}
$\mathbf{Z} \leftarrow [\,\mathbf{c}_{\mathrm{cls}} \,\|\,
  \{\mathbf{z}_i : a_i = 1\}\,]$ \tcp*{CLS never masked}
\For{$\ell \leftarrow 1$ \KwTo $L$}{
  $\mathbf{Z} \leftarrow \mathrm{Layer}_\ell(\mathbf{Z}, \mathbf{B})$
    \tcp*{fusion $h$, Eq.~\eqref{eq:layer},~\eqref{eq:mhsa}}
}
$\hat{\mathbf{p}} \leftarrow
  \mathrm{MLP}\big(\mathrm{CrossAttn}(\mathbf{q} + \phi(\tau), \mathbf{Z},
  \mathbf{B})\big)$ \tcp*{readout of $h$, Eq.~\eqref{eq:readout}}
$\mathcal{L} \leftarrow \mathcal{H}_\delta(\hat{\mathbf{p}} - \mathbf{p})$
  \tcp*{training, Eq.~\eqref{eq:loss}}
\Return $\hat{\mathbf{p}}$\;
\end{algorithm}

The fusion trunk uses $L=6$ pre-norm layers with $H=4$ heads. The complete model
holds about $1.5$~M parameters, roughly $0.05$~M per encoder.

\section{\uppercase{Experimental setup}}\label{sec:experiments}

\subsection{Datasets}

\noindent
The evaluation uses five datasets, one simulated and four real, each in a distinct
role; Table~\ref{tab:datasets} lists their settings and splits.

The Webots TIAGO++ simulation supplies co-recorded WiFi and IMU trajectories. Its WiFi field is synthesized rather than measured, so the accuracy
below one meter it permits is optimistic relative to a real deployment. MSILN
site1/B1~\cite{microsoft2021indoorloc} is the real cross-session test: its
training, validation and test paths are recorded on separate days, so the test
fingerprints belong to sessions the model never observed. IMUWiFine
floor~4~\cite{nurpeiissov2022imuwifine} holds out paths within a single session,
testing within-session behavior on real WiFi--IMU data.
UJIIndoorLoc~\cite{torressospedra2014ujiindoorloc} is a WiFi fingerprinting
benchmark of unordered scans, testing the WiFi encoder in isolation. RoNIN
canonical~\cite{yan2019ronin} holds out test sequences from subjects absent from
training, testing the IMU encoder in isolation.

\begin{table*}[tb]
\caption{The five datasets. The Unit column names what each split count counts,
  as the datasets are not structured alike; \#APs is the number of WiFi access
  points where the benchmark fixes it. MSILN is a Microsoft Indoor Localisation
  collection; Webots is a TIAGO++ simulation.}\label{tab:datasets}
\centering
\begin{tabular}{|l|l|l|c|c|c|c|c|}\hline
Dataset & Modalities & Setting & \#APs & Unit & Train & Val & Test \\ \hline
Webots sim & WiFi, IMU & Simulation (lab) & 117 & paths & 11 & 3 & 3 \\ \hline
MSILN site1/B1 & WiFi, IMU & Real (cross-session) & 1419 & paths & 94 & 34 & 5 \\ \hline
IMUWiFine fl.4 & WiFi, IMU & Real (held-out paths) & 343 & paths & 40 & 20 & 20 \\ \hline
UJIIndoorLoc & WiFi & Real (isolation) & 520 & scans & 19937 & 1111 & n/a \\ \hline
RoNIN canonical & IMU & Real (isolation) & n/a & sequences & 73 & n/a & 32 \\ \hline
\end{tabular}
\end{table*}

\subsection{Reference Methods}

\noindent
One reference method is reported per claim, and each table column names it.

\begin{itemize}
\item \textbf{wlan\_localization}~\cite{wlanloc}: $k$-nearest-neighbor
  ($k$-NN) RSSI fingerprinting ($k=3$, distance-weighted); the WiFi reference for
  both the encoder-only and the end-to-end fusion comparison.
\item \textbf{ResNet1D}~\cite{yan2019ronin}: a $4.6$~M-parameter learned regressor
  over inertial windows; the IMU reference for the encoder-only inertial
  comparison.
\item \textbf{PDR-from-start}: IMU-only pedestrian dead reckoning with step
  detection, anchored at the first ground-truth waypoint; the classical inertial
  reference for the fusion comparison.
\item \textbf{IMUWiFine}~\cite{nurpeiissov2022imuwifine}: a four-layer LSTM
  that fuses WiFi and IMU; the learned-fusion reference for the comparison across
  sessions.
\end{itemize}

\subsection{Metrics}

\noindent
Two metrics are reported, both in meters. Let $\hat{p}_i \in \mathbb{R}^2$ be the
predicted planar position at sample $i$, $p_i \in \mathbb{R}^2$ the ground truth,
and $N_s$ the number of samples. Mean absolute error (MAE) is the mean Euclidean
position error,
\begin{equation}
  \mathrm{MAE} = \frac{1}{N_s} \sum_{i=1}^{N_s} \lVert \hat{p}_i - p_i \rVert_2 ,
  \label{eq:mae}
\end{equation}
and absolute trajectory error (ATE) is the root-mean-square Euclidean position
error, reported without alignment as the raw ATE,
\begin{equation}
  \mathrm{ATE} = \sqrt{ \frac{1}{N_s} \sum_{i=1}^{N_s}
    \lVert \hat{p}_i - p_i \rVert_2^2 } .
  \label{eq:ate}
\end{equation}
Umeyama-aligned ATE~\cite{umeyama1991}, which removes a rigid-body offset by
similarity alignment before measuring error, appears only in the limitations
discussion as contextual framing for the inertial encoder.

\subsection{Implementation and Training}

\noindent
The set transformer of Section~\ref{sec:method} is trained for $40$ epochs
(AdamW, learning rate $1.3\times10^{-3}$, weight decay $10^{-4}$; OneCycleLR;
Huber loss with $\delta=0.5$; gradient clipping at $1.0$; batch size $128$; $K=4$), with
a fixed seed of $42$, in PyTorch on a single NVIDIA Quadro P4000.

\section{\uppercase{Experimental results}}\label{sec:results}

\subsection{Encoder Evaluation by Modality}

\noindent
Each encoder is first measured against its field's standard benchmark, so any
later fusion result rests on encoders already tested in isolation.
Table~\ref{tab:perleg} collects both comparisons.

On UJIIndoorLoc, WiFi-Net reaches an MAE of $8.69$~m on the validation scans,
against $15.17$~m for wlan\_localization\footnote{The $7.9$~m error in Section~\ref{sec:related} is the original 1-NN result on UJIIndoorLoc; wlan\_localization is an open-source $k$-nearest-neighbor implementation ($k=3$, distance-weighted) evaluated here on the validation split.}, a reduction of $42.7\%$. The benchmark
fixes $520$ access points and carries no trajectory structure, so the comparison
isolates the mapping from fingerprint to position with no temporal context.

\begin{figure}[tb]\centering
  \includegraphics[width=1\columnwidth]{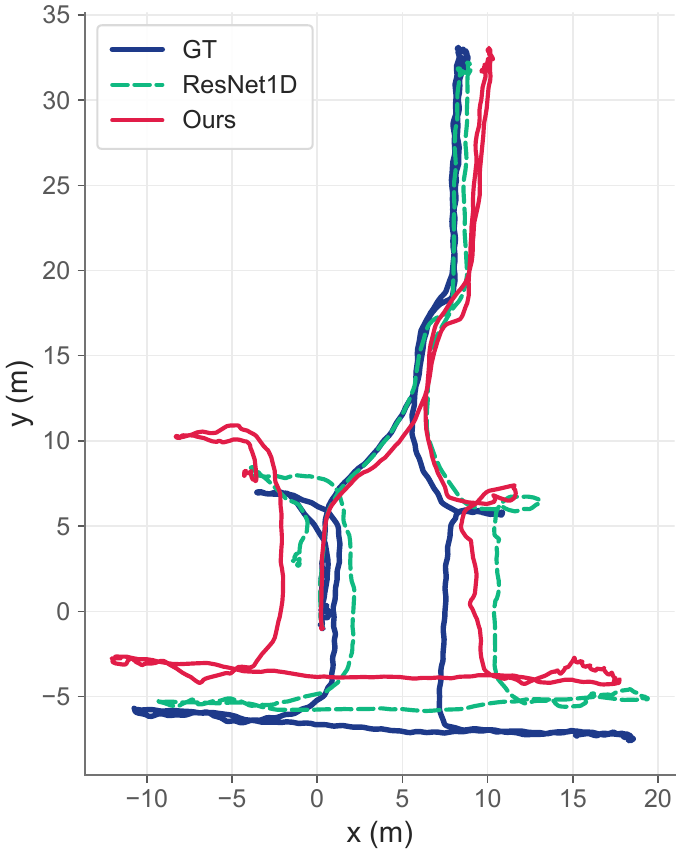}
  \caption{RoNIN canonical held-out sequence: ground truth, ResNet1D, and
    IMU-CNN (ours). The lighter inertial encoder follows the route but
    accumulates more drift than ResNet1D.}\label{fig:ronin_traj}
\end{figure}

The inertial comparison is more demanding. On RoNIN canonical, IMU-CNN attains a
raw ATE of $9.72$~m on the $32$ sequences held out from training, against
$5.14$~m for ResNet1D, trailing by $89.2\%$. The encoders differ sharply in size:
IMU-CNN holds about $0.05$~M parameters against ResNet1D's $4.6$~M, a factor of
roughly $95$. IMU-CNN is competitive within its own domain but pays a cross-subject
generalization cost, which it incurs on a significantly smaller budget. The drift is visible
on one such sequence in Figure~\ref{fig:ronin_traj}.

\begin{figure}[tb]\centering
  \includegraphics[width=1\columnwidth]{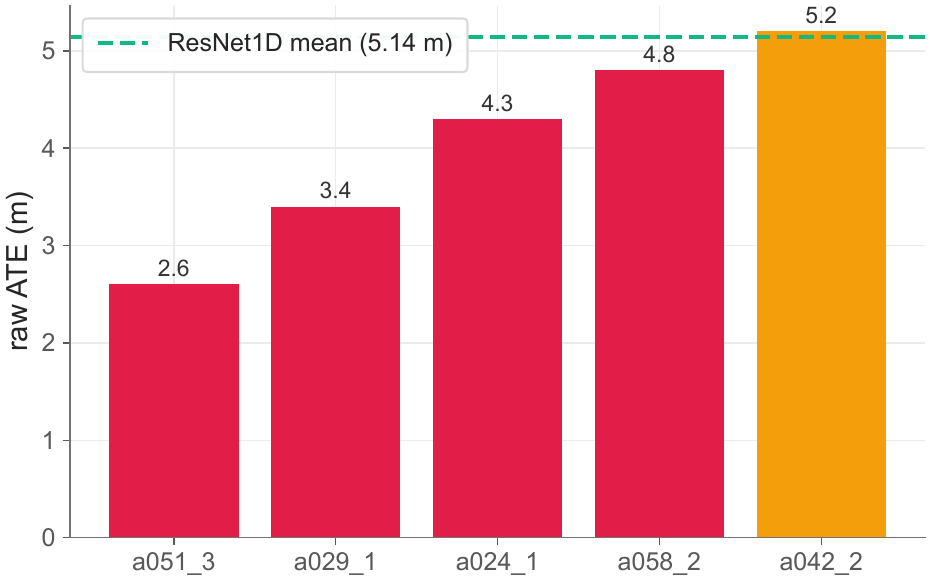}
  \caption{IMU-CNN's five best RoNIN canonical sequences by raw ATE, against the
    dashed ResNet1D mean of $5.14$~m. Four of the five fall below that mean, so
    the encoder is competitive per sequence despite trailing on the aggregate.}
  \label{fig:ronin_perseq}
\end{figure}

The aggregate figure hides a more even picture sequence by sequence.
Figure~\ref{fig:ronin_perseq} ranks IMU-CNN's five best sequences by raw
ATE against the ResNet1D mean of $5.14$~m: four fall below that mean. The $89.2\%$
gap thus comes from a few harder sequences, not a uniform shortfall.

\begin{table}[tb]
\caption{Encoder evaluation by modality, each encoder scored on its field's
  standard benchmark (MAE on UJIIndoorLoc, raw ATE on RoNIN canonical); the
  reference is wlan\_localization for UJIIndoorLoc and ResNet1D for RoNIN
  canonical. Bold marks the lowest error in each row.}\label{tab:perleg}
\centering
\footnotesize
\setlength{\tabcolsep}{2pt}
\begin{tabular}{|l|l|c|c|c|}\hline
Benchmark & Metric & Reference & Ours & Rel.\ $\Delta$ \\ \hline
UJIIndoorLoc (val) & MAE & 15.17 & \textbf{8.69} & $-42.7\%$ \\ \hline
RoNIN canonical & raw ATE & \textbf{5.14} & 9.72 & $+89.2\%$ \\ \hline
\end{tabular}
\end{table}

\subsection{End-to-End Fusion}

\begin{figure*}[t!]\centering
  \begin{subfigure}[t]{0.45\textwidth}\centering
    \includegraphics[width=\linewidth]{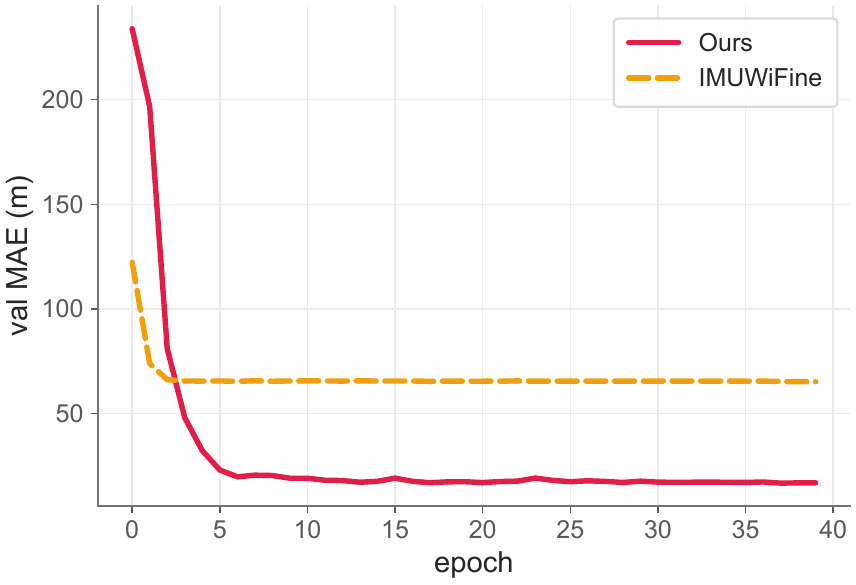}
    \caption{Validation MAE over training. The proposed model converges toward
      roughly $16$~m, while the IMUWiFine LSTM plateaus near $65$~m.}
    \label{fig:msiln_curves}
  \end{subfigure}\hfill
  \begin{subfigure}[t]{0.45\textwidth}\centering
    \includegraphics[width=\linewidth]{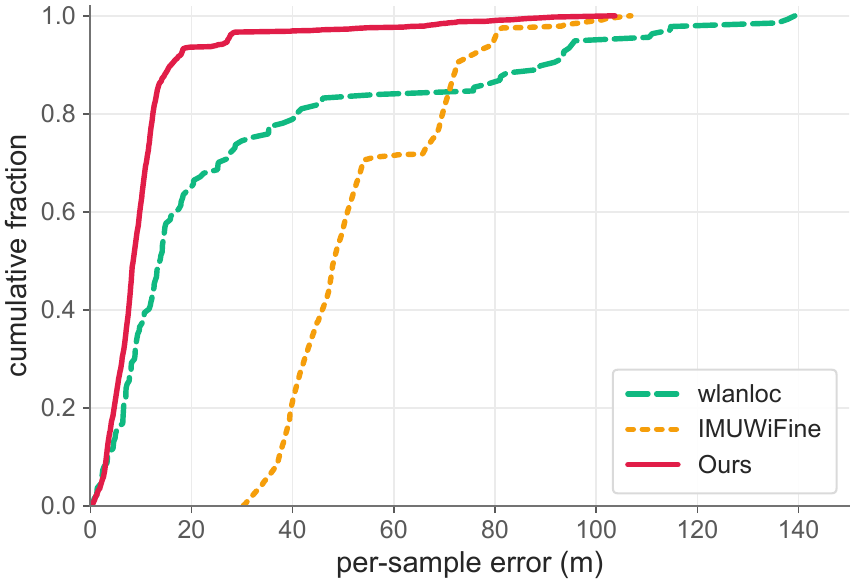}
    \caption{Error distribution per sample on the test session. The proposed
      method reaches $90\%$ of samples within roughly $15$~m.}
    \label{fig:msiln_cdf}
  \end{subfigure}\hfill
  \begin{subfigure}[t]{0.45\textwidth}\centering
    \includegraphics[width=\linewidth]{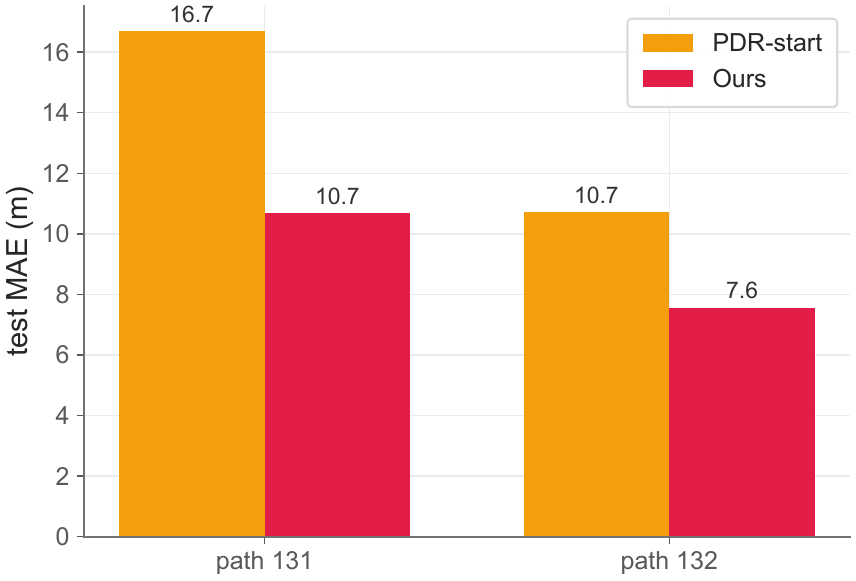}
    \caption{Test MAE for each of the two paths where the proposed method is ahead
      of PDR-from-start.}
    \label{fig:msiln_perpath}
  \end{subfigure}
  \caption{MSILN site1/B1 fusion across sessions. (\subref{fig:msiln_curves})
    training convergence against the IMUWiFine LSTM, (\subref{fig:msiln_cdf})
    test error distribution per sample against wlan\_localization and
    IMUWiFine, and (\subref{fig:msiln_perpath}) MAE per path against
    PDR-from-start on the paths where the proposed method is ahead.}
  \label{fig:msiln}
\end{figure*}

\noindent
With both encoders validated, the set transformer fuses them, evaluated first in
the controlled simulation, then on the real collection across sessions that
carries the headline claim. Table~\ref{tab:fusion} collects the fusion results
against the three named reference methods.

\begin{figure}[!hb]\centering
  \includegraphics[width=1\columnwidth]{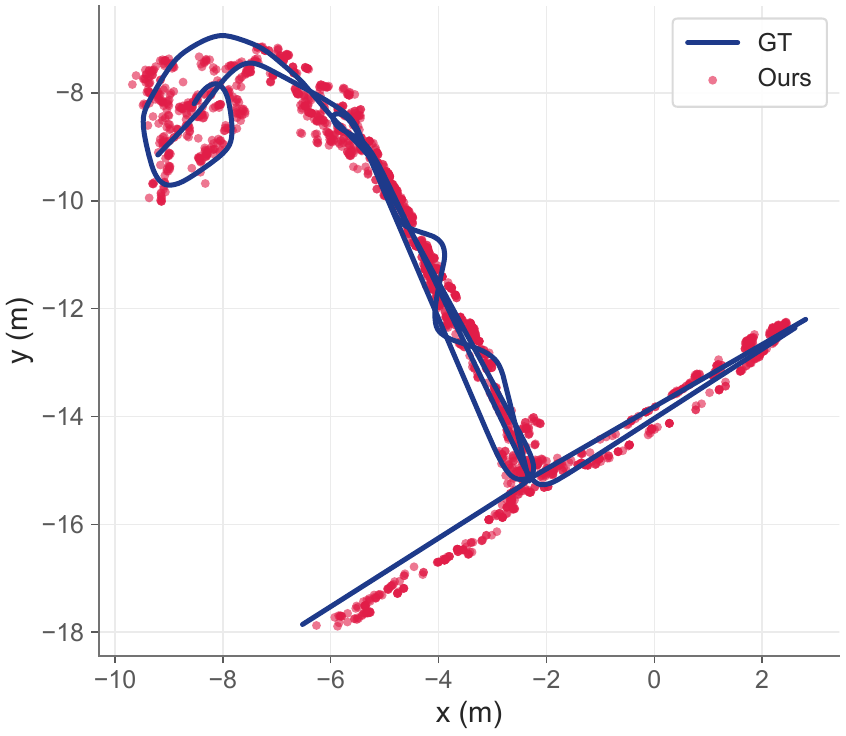}
  \caption{Webots simulation test paths: predicted positions (ours) against
    ground truth. The controlled lab admits tracking below one meter when the WiFi
    field is clean.}\label{fig:webots_scatter}
\end{figure}

In the Webots simulation the WiFi field is clean by construction. There the
model fusing the two modalities at $K=4$ reaches $0.61$~m MAE on test
(Figure~\ref{fig:webots_scatter}). This figure, below one meter, is a controlled
reference point, not a deployment claim: with a synthesized WiFi field it is not
directly comparable to the real numbers below.

MSILN site1/B1 supplies the real evaluation across sessions. On the test session,
held out for evaluation, the set transformer reaches $10.90$~m MAE, against
$28.31$~m for wlan\_localization, $12.49$~m for PDR-from-start and $52.69$~m for
IMUWiFine. On the
validation session the margin is narrow: $16.67$~m against $16.88$~m for
PDR-from-start, with wlan\_localization at $21.26$~m and IMUWiFine at $65.15$~m.
Where wlan\_localization nearly doubles its error from validation to test, the
fusion error falls. Fusing the dense inertial stream against the sparse absolute
reference maintains position across the session boundary. IMUWiFine, itself a learned
WiFi--IMU fusion, remains above $52$~m on both splits
(Figure~\ref{fig:msiln_curves}), an informative negative result for learned fusion
under fingerprint drift. Beating wlan\_localization, a distance-weighted WiFi
$k$-NN with no motion model, is expected of any method that tracks motion, so the
$-61.5\%$ figure is a sanity check; the comparisons that carry weight are against
PDR-from-start and the IMUWiFine LSTM. The full distribution of test error per
sample is in Figure~\ref{fig:msiln_cdf}, and error for each path against
PDR-from-start on the two paths where the proposed method is ahead in
Figure~\ref{fig:msiln_perpath}.

\begin{table*}[tb]
\caption{End-to-end fusion: position MAE (m). The simulation has no public
  reference fusing two modalities; the MSILN comparison runs across sessions
  against three named reference methods. Bold marks the lowest error in each
  row.}\label{tab:fusion}
\centering
\begin{tabular}{|l|c|c|c|c|}\hline
Dataset (split) & wlan\_localization & PDR-from-start & IMUWiFine & Ours \\ \hline
Webots sim (test) & n/a & n/a & n/a & \textbf{0.61} \\ \hline
MSILN site1/B1 (val) & 21.26 & 16.88 & 65.15 & \textbf{16.67} \\ \hline
MSILN site1/B1 (test) & 28.31 & 12.49 & 52.69 & \textbf{10.90} \\ \hline
\end{tabular}
\end{table*}

\subsection{Generalization to a Further Dataset}

\noindent
IMUWiFine floor~4~\cite{nurpeiissov2022imuwifine} provides the complementary test:
a split that holds out paths within the same distribution, on the LSTM's own
dataset, where the proposed method gains nothing from session drift. It reaches a
test MAE of $6.37$~m on this split, against a validation MAE of $2.31$~m.

\begin{figure}[!hb]\centering
  \includegraphics[width=1\columnwidth]{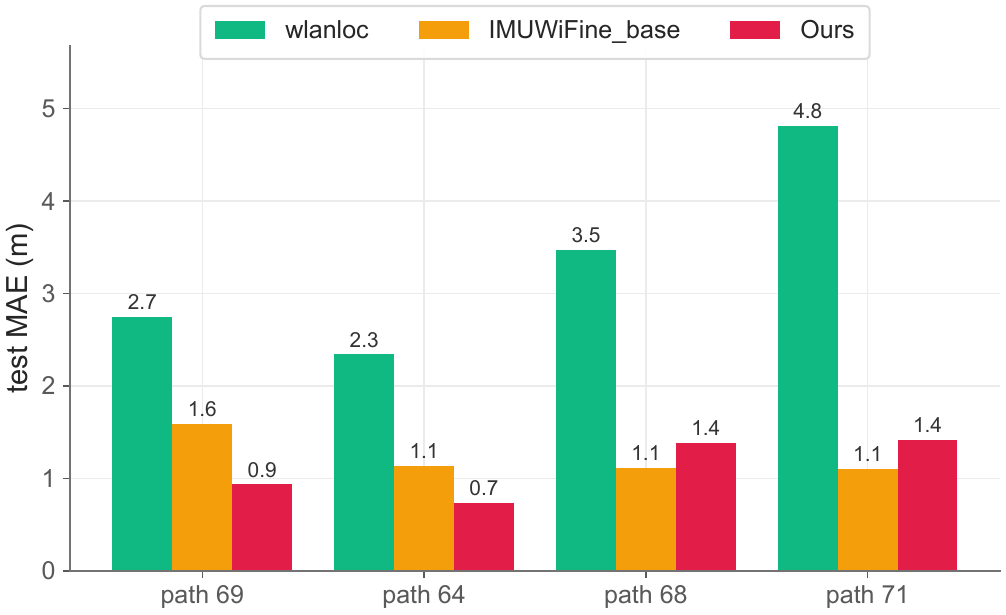}
  \caption{IMUWiFine floor~4: test MAE (m) for each of four paths where the proposed
    method is best or competitive.}\label{fig:iwfine_showcase}
\end{figure}

On four paths where the proposed method is best or competitive
(Figure~\ref{fig:iwfine_showcase}), the LSTM tracks its home ground to roughly one
meter, and the proposed method takes two of them (path~$69$ at $0.93$~m and
path~$64$ at $0.73$~m) while trailing narrowly on the others. Both learned methods
improve over wlan\_localization by roughly two to three times on every path. The
LSTM is therefore strong within its own distribution but not across sessions,
whereas the proposed method is competitive within distribution and holds up on the
MSILN test across sessions.

\subsection{Robustness and Ablations}

\noindent
The remaining experiments run on the Webots simulation, where every signal is
controlled and individual sensors can be suppressed or aged on demand, realizing
the missing ($a_i=0$) and stale (large $|\Delta t_i|$) conditions defined in
Section~\ref{sec:problem}.

\textbf{Modality dropout.} Suppressing a modality at test time exposes what each
stream contributes. With WiFi only the test MAE is $0.86$~m; with IMU only it is
$3.50$~m; with both it is $0.61$~m. The WiFi token provides the absolute
reference, so dropping IMU costs little, while IMU alone drifts without it.

\textbf{Staleness.} Aging the WiFi token while the IMU stream stays fresh exposes
the cost of delayed measurements. The test MAE climbs gradually,
$0.61 \to 1.13 \to 1.48 \to 1.84 \to 2.05$~m, reaching the level of IMU alone at
$3.50$~m only once the reference is fully stale: a slope rather than a cliff across
moderate delays.

\textbf{Latency.} On the Quadro P4000 the model runs at $4.77$~ms per sample at
batch size $1$ and $0.146$~ms at batch size $32$, fast enough for real-time use.

\subsection{Limitations}

\noindent
Two limitations qualify the results. First, the $+89.2\%$ inertial gap on RoNIN
canonical persists under Umeyama alignment~\cite{umeyama1991}, narrowing only to
$+48.2\%$ ($7.62$~m versus ResNet1D's $5.14$~m).
Second, the MSILN test split is dominated by one WiFi dense path of $786$ samples,
roughly $28\%$ of the test mass, so the test figure is best read alongside the
validation figure rather than in isolation.

\section{\uppercase{Conclusion}}\label{sec:conclusion}

\noindent We presented in this paper a continuous-time Set-Transformer for asynchronous WiFi-IMU indoor localization. Each observation is represented as a token with a learned encoding of its elapsed time, and a single permutation-invariant self-attention block fuses the resulting set. Modality and instant dropout make robustness to missing and stale observations a property of the training procedure rather than the architecture itself. On standard single-modality benchmarks, the WiFi encoder outperforms an open-source fingerprinting baseline, while the lightweight inertial encoder remains competitive with a significantly larger reference model despite using fewer parameters. The fused model also outperforms an open WiFi reference method under real cross-session drift, the setting most representative of practical deployment, while degrading gracefully in the presence of missing or stale inputs.

The approach has several limitations. The simulated WiFi environment is synthetic, making its sub-meter accuracy optimistic. The inertial encoder still underperforms a significantly larger reference model on cross-subject trajectories, and the cross-session margin is influenced, in part, by the composition of the test split. Future work will focus on integrating additional sensing modalities to improve robustness, as well as incorporating physical constraints into both the fusion mechanism and the training process.

\ifblind\else
\section*{\uppercase{Acknowledgements}}
\noindent This work is funded in part by \textit{Région Grand Est (RGE), France}.
\fi 

\bibliographystyle{apalike}
{\small \bibliography{refs}}

@MISC{wlanloc,
  author = {Naribole, Sharan},
  title = {{wlan\_localization}: {WiFi} {RSSI} Fingerprinting for Indoor Localization},
  year = {2016},
  howpublished = {GitHub: github.com/sharan-naribole/wlan\_localization},
  note = {Open-source implementation, WiFi baseline of record}
}

@MISC{nurpeiissov2022imuwifine,
  author = {Nurpeiissov, Mukhamet and Kuzdeuov, Askat and Assylkhanov, Aslan and Khassanov, Yerbolat and Varol, Atakan},
  title = {End-to-End Sequential Indoor Localization Using Smartphone Inertial Sensors and {WiFi}},
  year = {2022},
  howpublished = {TechRxiv preprint},
  doi = {10.36227/techrxiv.17109275.v1},
  note = {Introduces the {IMUWiFine} dataset; learned {WiFi}+{IMU} fusion baseline}
}

@INPROCEEDINGS{bahl2000radar,
  author = {Bahl, Paramvir and Padmanabhan, Venkata N.},
  title = {{RADAR}: An In-Building {RF}-Based User Location and Tracking System},
  booktitle = {Proceedings IEEE INFOCOM 2000. Conference on Computer Communications. Nineteenth Annual Joint Conference of the IEEE Computer and Communications Societies},
  year = {2000},
  volume = {2},
  pages = {775--784},
  doi = {10.1109/INFCOM.2000.832252}
}

@INPROCEEDINGS{torressospedra2014ujiindoorloc,
  author = {Torres-Sospedra, Joaqu\'{i}n and Montoliu, Raul and Mart\'{i}nez-Us\'{o}, Adolfo and Avariento, Joan and Arnau, Tomas and Benedito-Bordonau, Mauri and Huerta, Joaqu\'{i}n},
  title = {{UJIIndoorLoc}: A New Multi-Building and Multi-Floor Database for {WLAN} Fingerprint-Based Indoor Localization Problems},
  booktitle = {2014 International Conference on Indoor Positioning and Indoor Navigation (IPIN)},
  year = {2014},
  doi = {10.1109/IPIN.2014.7275492}
}

@INPROCEEDINGS{nguyen2025imot,
  author = {Nguyen, Son Minh and Le, Duc Viet and Havinga, Paul J. M.},
  title = {{iMoT}: Inertial Motion {Transformer} for Inertial Navigation},
  booktitle = {Proceedings of the {AAAI} Conference on Artificial Intelligence},
  year = {2025},
  volume = {39},
  number = {6},
  pages = {6209--6217},
  doi = {10.1609/aaai.v39i6.32664}
}

@INPROCEEDINGS{yan2019ronin,
  author = {Herath, Sachini and Yan, Hang and Furukawa, Yasutaka},
  title = {{RoNIN}: Robust Neural Inertial Navigation in the Wild: Benchmark, Evaluations, and New Methods},
  booktitle = {{IEEE} International Conference on Robotics and Automation ({ICRA})},
  year = {2020},
  pages = {3146--3152},
  doi = {10.1109/ICRA40945.2020.9196860},
  publisher = {IEEE}
}

@ARTICLE{zhou2024wioekf,
  author = {Zhou, Pan and Wang, Hao and Gravina, Raffaele and Sun, Fangmin},
  title = {{WIO-EKF}: Extended {Kalman} Filtering-Based {Wi-Fi} and Inertial Odometry Fusion Method for Indoor Localization},
  journal = {IEEE Internet of Things Journal},
  year = {2024},
  volume = {11},
  number = {13},
  pages = {23592--23603},
  doi = {10.1109/JIOT.2024.3386889}
}

@ARTICLE{wei2021sensorfusion,
  author = {Wei, Xijia and Wei, Zhiqiang and Radu, Valentin},
  title = {Sensor-Fusion for Smartphone Location Tracking Using Hybrid Multimodal Deep Neural Networks},
  journal = {Sensors},
  year = {2021},
  volume = {21},
  number = {22},
  pages = {7488},
  doi = {10.3390/s21227488}
}

@ARTICLE{neverova2014moddrop,
  author = {Neverova, Natalia and Wolf, Christian and Taylor, Graham and Nebout, Florian},
  title = {{ModDrop}: Adaptive Multi-Modal Gesture Recognition},
  journal = {IEEE Transactions on Pattern Analysis and Machine Intelligence},
  year = {2016},
  volume = {38},
  doi = {10.1109/TPAMI.2015.2461544}
}

@INPROCEEDINGS{vaswani2017transformer,
  author = {Vaswani, Ashish and Shazeer, Noam and Parmar, Niki and Uszkoreit, Jakob and Jones, Llion and Gomez, Aidan N. and Kaiser, {\L}ukasz and Polosukhin, Illia},
  title = {Attention is All You Need},
  booktitle = {Advances in Neural Information Processing Systems (NeurIPS)},
  year = {2017},
  volume = {30},
  publisher = {Curran Associates, Inc.},
  url = {https://proceedings.neurips.cc/paper_files/paper/2017/hash/3f5ee243547dee91fbd053c1c4a845aa-Abstract.html}
}

@INPROCEEDINGS{lee2019settransformer,
  author = {Lee, Juho and Lee, Yoonho and Kim, Jungtaek and Kosiorek, Adam and Choi, Seungjin and Teh, Yee Whye},
  title = {Set {Transformer}: A Framework for Attention-based Permutation-Invariant Neural Networks},
  booktitle = {Proceedings of the 36th International Conference on Machine Learning (ICML)},
  year = {2019},
  publisher = {PMLR},
  url = {https://proceedings.mlr.press/v97/lee19d.html}
}

@INPROCEEDINGS{kaygusuz2022aftvo,
  author = {Kaygusuz, Nimet and Mendez, Oscar and Bowden, Richard},
  title = {{AFT-VO}: Asynchronous Fusion Transformers for Multi-View Visual Odometry Estimation},
  booktitle = {{IEEE/RSJ} International Conference on Intelligent Robots and Systems ({IROS})},
  year = {2022},
  pages = {2402--2408},
  publisher = {IEEE}
}

@INPROCEEDINGS{shukla2021mtan,
  author = {Shukla, Satya Narayan and Marlin, Benjamin M.},
  title = {Multi-Time Attention Networks for Irregularly Sampled Time Series},
  booktitle = {International Conference on Learning Representations ({ICLR})},
  year = {2021}
}

@MISC{kazemi2019time2vec,
  author = {Kazemi, Seyed Mehran and Goel, Rishab and Eghbali, Sepehr and Ramanan, Janahan and Sahota, Jaspreet and Thakur, Sanjay and Wu, Stella and Smyth, Cathal and Poupart, Pascal and Brubaker, Marcus},
  title = {{Time2Vec}: Learning a Vector Representation of Time},
  year = {2019},
  howpublished = {arXiv:1907.05321},
  doi = {10.48550/arXiv.1907.05321}
}

@INPROCEEDINGS{chen2023contiformer,
  author = {Chen, Yuqi and Ren, Kan and Wang, Yansen and Fang, Yuchen and Sun, Weiwei and Li, Dongsheng},
  title = {{ContiFormer}: Continuous-Time {Transformer} for Irregular Time Series Modeling},
  booktitle = {Advances in Neural Information Processing Systems ({NeurIPS})},
  year = {2023}
}

@INPROCEEDINGS{horn2020seft,
  author = {Horn, Max and Moor, Michael and Bock, Christian and Rieck, Bastian and Borgwardt, Karsten},
  title = {Set Functions for Time Series},
  booktitle = {Proceedings of the 37th International Conference on Machine Learning ({ICML})},
  year = {2020},
  publisher = {PMLR}
}

@ARTICLE{tipirneni2022strats,
  author = {Tipirneni, Sindhu and Reddy, Chandan K.},
  title = {Self-Supervised {Transformer} for Sparse and Irregularly Sampled Multivariate Clinical Time-Series},
  journal = {ACM Transactions on Knowledge Discovery from Data},
  year = {2022},
  volume = {16},
  number = {6},
  pages = {105:1--105:17},
  doi = {10.1145/3516367}
}

@ARTICLE{umeyama1991,
  author = {Umeyama, Shinji},
  title = {Least-Squares Estimation of Transformation Parameters Between Two Point Patterns},
  journal = {IEEE Transactions on Pattern Analysis and Machine Intelligence},
  year = {1991},
  volume = {13},
  number = {4},
  pages = {376--380},
  doi = {10.1109/34.88573}
}

@MISC{microsoft2021indoorloc,
  author = {{Microsoft Research}},
  title = {Indoor Location \& Navigation},
  year = {2021},
  howpublished = {Kaggle competition},
  note = {\url{https://www.kaggle.com/competitions/indoor-location-navigation}}
}

\end{document}